\pdfoutput=1

\documentclass[11pt]{article}

\usepackage{emnlp2021}

\usepackage{times}
\usepackage{latexsym}

\usepackage[T1]{fontenc}

\usepackage[utf8]{inputenc}

\usepackage{microtype}

\usepackage{multirow}
\usepackage{graphicx}
\usepackage{subcaption}
\usepackage{adjustbox}
\usepackage{amsmath}
\usepackage{amsfonts}
\usepackage{threeparttable}
\usepackage{booktabs}
\usepackage{enumitem}
\usepackage{stfloats}

\newcommand{\xhdr}[1]{{\noindent\bfseries #1}.}
\newcommand{\squishlist}{
 \begin{list}{$\bullet$}
  { \setlength{\itemsep}{0pt}
     \setlength{\parsep}{1pt}
     \setlength{\topsep}{1pt}
     \setlength{\partopsep}{0pt}
     \setlength{\leftmargin}{1em}
     \setlength{\labelwidth}{1em}
     \setlength{\labelsep}{0.5em} } }
\newcommand{\squishend}{\end{list}}

\title{BERT4GCN: Using BERT Intermediate Layers to Augment GCN for Aspect-based Sentiment Classification}
\author{Zeguan Xiao\textsuperscript{1}, 
Jiarun Wu\textsuperscript{1}, 
Qingliang Chen\textsuperscript{1,2}\thanks{\quad Corresponding author (Qingliang Chen).},
Congjian Deng\textsuperscript{3} \\
\textsuperscript{1}Department of Computer Science, Jinan University, Guangzhou 510632, China \\
\textsuperscript{2}Guangzhou Xuanyuan Research Institute Company, Ltd., Guangzhou 510006, China\\
\textsuperscript{3}Guangzhou Yunqu Information Technology Company, Ltd., Guangzhou 510665, China \\

\texttt{xiaozg@stu2018.jnu.edu.cn},~
\texttt{benwu@stu2019.edu.cn} \\
\texttt{tpchen@jnu.edu.cn},~
\texttt{dcj@yunqu-info.com} \\
}

\begin{document}
\maketitle

\begin{abstract}
Graph-based Aspect-based Sentiment Classification (ABSC) approaches have yielded state-of-the-art results, expecially when equipped with contextual word embedding from pre-training language models (PLMs). However, they ignore sequential features of the context and have not yet made the best of PLMs. In this paper, we propose a novel model, BERT4GCN, which integrates the grammatical sequential features from the PLM of BERT, and the syntactic knowledge from dependency graphs. BERT4GCN utilizes outputs from intermediate layers of BERT and positional information between words to augment GCN (Graph Convolutional Network) to better encode the dependency graphs for the downstream classification. Experimental results demonstrate that the proposed BERT4GCN outperforms all state-of-the-art baselines, justifying that augmenting GCN with the grammatical features from intermediate layers of BERT can significantly empower ABSC models.
\end{abstract}
\section{Introduction}
Aspect-based sentiment classification (ABSC), a fine-grained sentiment classification task in the field of sentiment analysis, aims at identifying the sentiment polarities of aspects explicitly given in sentences. For example, given the sentence \emph{``The price is reasonable although the service is poor.''}, ABSC needs to correctly assign a positive polarity to \emph{price} and a negative one to \emph{service}.

Intuitively, matching aspects with their corresponding opinion expressions is the core of ABSC task. Some of previous deep learning approaches \citep{tang-etal-2016-aspect,ma2017interactive,huang2018aspect,song2019attentional} use various types of attention mechanisms to model the relationship between aspects and opinion expressions in an implicit way. However, these attention-based models do not take good advantage of the syntactic information of sentences, such as the dependency graph, to better associate the aspects with their sentimental polarities, thus leading to poor performance.

In order to better pair aspects and their corresponding opinion expressions, the syntactic features must be taken into account. Recent work such as \citep{huang2019syntax,sun2019aspect,zhang2019aspect,wang2020relational,tang-etal-2020-dependency} employs either Graph Convolutional Networks (GCNs) \citep{kipf2016semi} or Graph Attention Networks (GATs) \citep{velivckovic2017graph} on the dependency graph, in which words in the sentence are nodes in the graph and grammatical relations are edge labels. The dependency graph they use can be obtained from either ordinary dependency tree parser or reshaping it using heuristic rules. These models can achieve promisingly better performance with this additional grammatical features, especially when they incorporate the contextual word embedding, such as BERT \citep{devlin2018bert}.

Meanwhile, BERTology researchers have investigated what linguistic knowledge can be learned from unlabeled data by language models  \citep{clark2019does,hewitt-liang-2019-designing,hewitt-manning-2019-structural,jawahar-etal-2019-bert}. And it has been shown that BERT captures a rich hierarchy of linguistic information, with surface features in lower layers, syntactic features in middle layers and semantic features in higher layers  \citep{jawahar-etal-2019-bert}.

Inspired by \citet{jawahar-etal-2019-bert}, we propose a novel framework of BERT4GCN to make the best of the rich hierarchy of linguistic information from BERT in this paper.
Specifically, we firstly encode the context with a BiLSTM (Bidirectional Long Short-Term Memory) to capture the contextual information regarding word orders. Then, we use the hidden states of BiLSTM to initiate node representations and employ multi-layer GCN on the dependency graph.
Next, we combine the node representations of each layer of GCN with the hidden states of some intermediate layer of BERT, i.e., for the neighborhood aggregation of GCN. In this way, BERT4GCN can fuse grammatical sequential features with graph-based representations. Besides, we further prune and add the edge of the dependency graph based on self-attention weights in the Transformer encoder \citep{vaswani2017attention} of BERT to deal with parsing errors and make dependency graph better suit ABSC task. In addition, we develop a method which incorporates relative positional embedding in node representations to make GCN position-aware.

Our contributions in this paper are summarized as follows:
{
\squishlist
\item  We propose a BERT4GCN model, which naturally incorporates grammatical sequential features and syntactic knowledge from intermediate layers in BERT to augment GCN and thus can produce better encodings for the downstream ABSC task. BERT4GCN can make the best of the hidden linguistic knowledge in BERT without learning from scratch on other sources. 
\item The experiments have been conducted on SemEval 2014 Task 4 \citep{pontiki-etal-2014-semeval} and Twitter \citep{dong-etal-2014-adaptive} datasets, with promising results showing that BERT4GCN achieves new state-of-the-art performance across these prestigious benchmarks.
\squishend
}
\section{Related Work}
Modeling the connection between aspect terms and opinion expressions is the core of the ABSC task. State-of-the-art models combine GNNs (Graph Neural Networks) with dependency graphs whose syntactic information is very helpful. Some have stacked several GNN layers to propagate sentiment from opinion expressions to aspect terms \citep{huang2019syntax, sun2019aspect, zhang2019aspect}. Some have tried to convert the original dependency graph to the aspect-oriented one and encode dependency relations \citep{wang2020relational}. More recently, a dual structure model has been proposed in \citep{tang-etal-2020-dependency}, jointly taking the sequential features and dependency graph knowledge together. Our BERT4GCN model has a similar structure to that in \citep{tang-etal-2020-dependency}, but ours utilizes the grammatical sequential features directly from intermediate layers in BERT, instead of learning from scratch on other sources.

\section{BERT4GCN}

\subsection{Word Embedding and BiLSTM}
Given $s = [w_{1},w_{2},...,w_{i},...,w_{i+m-1},...,w_{n}]$ as a sentence of $n$ words and a substring $a = [w_{i},...,w_{i+m-1}]$ representing aspect terms, we first map each word to a low-dimensional word vector. For each word $w_{i}$, we get a vector $ x_{i} \in \mathbb{R}^{d_{e}} $ where $d_{e}$ is the dimensionality of the word embedding. After that, we employ a BiLSTM to word embeddings to produce hidden state vectors $\mathbf{H} = [h_{1},h_{2},...,h_{n} ]$, where each $ h_{t} \in \mathbb{R}^{2d_{h}}$ represents the hidden state at time step $t$ from bidirectional LSTM, and $d_{h}$ is the dimensionality of a hidden state output by an unidirectional LSTM.
$\mathbf{H}$ will fuse with BERT hidden states to produce the input of the first GCN layer.

\subsection{The Usage of BERT}
We consider BERT as a grammatical knowledge encoder that encodes input text features in hidden states and self-attention weights. The input of BERT model is formulated as [CLS]$s$[SEP]$a$[SEP], where $s$ is the sentence and $a$ is the aspect term.

\subsubsection{Hidden States as Augmented Features}
BERT captures a rich hierarchy of linguistic information, spreading over the Transformer encoder blocks \citep{clark2019does, jawahar-etal-2019-bert}.
\citet{jawahar-etal-2019-bert} also show that neighbouring layers of BERT learn similar linguistic knowledge. Therefore we select layers uniformly from BERT as the source of augmented features.
BERT4GCN utilizes the hidden states of the 1st, 5th, 9th and 12th layers of BERT-Base model as augmented features\footnote{We also try other layers to build $\rm BERT4GCN_{2,7,12}$ and $\rm BERT4GCN_{1,4,7,10,12}$ (the numbers in subscript mean layers used), but the model in the paper is the best in experimental results on average.}. We define the hidden states of all BERT layers as $ \mathbf{B} = [\mathbf{H}^B_{1}, \mathbf{H}^B_{2},...,\mathbf{H}^B_{12}] $, where each $ \mathbf{H}^B_{i} \in \mathbb{R}^{n \times d_{BERT}}$, and $d_{BERT}$ is the dimensionality of the hidden state. When a word is split into multiple sub-words, we just use the hidden state corresponding to its first sub-word. Then we can get the augmented features $\mathbf{G}$ for GCN as:
\begin{align}
\mathbf{G} &= \left [ \mathbf{H}^B_{1},\mathbf{H}^B_{5}, \mathbf{H}^B_{9},\mathbf{H}^B_{12} \right ]
\end{align}

\subsubsection{Supplementing Dependency Graph}
Self-attention mechanism captures long-distance dependencies between words.
Therefore, we apply self-attention weights of BERT to supplement dependency graphs to deal with parsing errors and make dependency graphs better suit the ABSC task. After getting dependency graphs from an ordinary parser, we substitute the unidirectional edges to bidirectional ones.

Next, we obtain attention weight tensors $\mathbf{A}^{att} = \left [ \mathbf{W}_{1}^{att},\mathbf{W}_{5}^{att},\mathbf{W}_{9}^{att},\mathbf{W}_{12}^{att} \right ]$, where  each $\mathbf{W}_{i}^{att} \in \mathbb{R}^{h \times n \times n}$ is the self-attention weight tensor of $i$-th layer Transformer encoder of BERT, and $h$ is the number of attention heads.
Then, we average them over the head dimension $\bar{\mathbf{A}}_{l}^{att} = \frac{1}{h} \sum_{i=1}^{h} \mathbf{A}_{l,i}^{att}$, where $\mathbf{A}_{l}^{att}$ is the $l$-th element in $\mathbf{A}^{att}$, and $l \in \left \{ 1,2,3,4 \right \}$.
Finally, we prune or add directed edges between words if the attention weight is larger or smaller than some thresholds. So the supplemented dependency graph for the $l$-th layer of GCN is formulated as follows:
\begin{align}
\mathbf{A}_{l,i,j}^{sup} &= \left\{
\begin{array}{rcl}
1 & & {\alpha \leq \bar{\mathbf{A}}_{l,i,j}^{att}} \\
\mathbf{A}_{i,j} & & {\beta < \bar{\mathbf{A}}_{l,i,j}^{att} < \alpha} \\
0 & & {\bar{\mathbf{A}}_{l,i,j}^{att} \leq \beta}
\end{array} \right.
\end{align}
where $\mathbf{A}$ is the adjacency matrix of the original dependency graph, $\mathbf{A}_{l}^{sup}$ is the  adjacency matrix of the supplemented dependency graph and $\mathbf{A}_{l,i,j}^{sup}$ is the element on the $i$-th row and $j$-th column of $\mathbf{A}_{l}^{sup}$, $\alpha$ and $\beta$ are hyperparameters.

\subsection{GCN over Supplemented Dependency Graph}
We then apply GCN over the supplemented dependency graph in every layer, whose input is a fusion of BERT hidden states and output of the previous layer as follows:
\begin{align}
\mathbf{R}_{1} &= {\rm ReLU} \left ( \mathbf{G}_{1} \mathbf{W}_{1} \right ) + \mathbf{H} \\
\mathbf{R}_{k} &= {\rm ReLU}\left ( \mathbf{G}_{k} \mathbf{W}_{k} \right ) + \mathbf{O}_{k-1} \\
\mathbf{O}_{l,i} &= {\rm ReLU} \left ( \frac{1}{d_{i}} \sum_{j=1}^{n}
\mathbf{A}_{l,i,j}^{sup} \mathbf{W}^{l} \mathbf{R}_{l,j} + \mathbf{b}^{l} \right )
\end{align}
where $ k \in \left \{ 2,3,4 \right \}, l \in \left \{ 1,2,3, 4 \right \} $, $\mathbf{G}_{1}$ and $ \mathbf{G}_{k} \in \mathbb{R}^{n \times d_{BERT}} $ is the first and $k$-th element in $\mathbf{G}$, respectively. $\mathbf{R}_{1}$ and $\mathbf{R}_{k}$ are the node representations which fuse the hidden states of BERT with the hidden states of BiLSTM or the output of the preceding GCN layer, before feeding to the first and $k$-th GCN layer. $\mathbf{O}_{l,i}$ is output of $l$-th layer in GCN. $d_{i}$ is the outdegree of node $i$. The weight $\mathbf{W}_{1}$,$\mathbf{W}_{k},\mathbf{W}^{l}$ and bias $\mathbf{b}^{l}$ are trainable parameters.

\subsection{Taking Relative Positions}
The GCN aggregates neighbouring nodes representations in an averaging way and ignores the relative linear positions in the original context. To address this issue, we learn a set of relative position embeddings $\mathbf{P} \in \mathbb{R}^{(2w+1) \times d_{p}}$ to encode positional information, where $w$ is the positional window hyperparameter. Before aggregating neighbouring node representation, we add the relative position embedding to the node representation, formalizing  the relative linear position to the current word as follows:
\begin{gather}
\mathbf{O}_{l,i} = {\rm ReLU} \left ( \frac{1}{d_{i}} \sum_{j=1}^{n} \mathbf{A}_{l,i,j}^{sup} \mathbf{W}^{l} \mathbf{R}_{l,j}^{p} + \mathbf{b}^{l} \right ) \\
\mathbf{R}_{l,j}^{p} = \mathbf{R}_{l,j} + \mathbf{P}_{clip\left ( j-i,w \right )} \\
clip\left ( x,w \right ) = max \left ( -w, min(w,x) \right )
\end{gather}
where $\mathbf{R}_{l,j}^{p}$ is the positional node representation, and $clip$ function returns the embedding index.

\subsection{The Training}
Having obtained the representation of words after the last GCN layer, we average the representation of the current aspect terms as the final feature for classification:
\begin{align}
h_{a} &= \frac{1}{m} \sum_{p=i}^{i+m-1} \mathbf{O}_{4,p}
\end{align}
It is then fed into a fully-connected layer, followed by a softmax layer to yield a probability distribution $ \mathbf{p}$ over polarity space:
\begin{align}
\mathbf{p} &= softmax\left ( h_{a} \mathbf{W}_{c} + \mathbf{b}_{c} \right )
\end{align}
where $\mathbf{W}_{c}$ and $\mathbf{b}_{c}$ are trainable weights and biases, respectively.

The proposed BERT4GCN is optimized by the gradient descent algorithm with the cross entropy loss and $L_2$-regularization:
\begin{align}
\rm Loss &= - \sum_{\left ( x,y \right ) \in \mathbb{D}} \mathbf{ln} \mathbf{p}_{y} + \lambda {\left \| \Theta  \right \|}_{2}
\end{align}
where $\mathbb{D}$ denotes the training dataset, $y$ is the ground-truth label, $\mathbf{p}_{y}$ is the $y$-th element of $\mathbf{p}$, $\Theta$ represents all trainable parameters, and $\lambda$  is the coefficient of the regularization term.
\section{Experiments}

\subsection{Datasets and Experimental Setup}
We have evaluated our model on three widely used datasets, including Laptop and Restaurant datasets from SemEval 2014 Task 4 \citep{pontiki-etal-2014-semeval} and Twitter datasets from \citet{dong-etal-2014-adaptive}.  And we compare the proposed BERT4GCN with a series of baselines and state-of-the-art models, including LSTM, BERT-SPC \citep{song2019attentional}, RoBERTa-MLP \citep{dai2021does}, MemNet \citep{tang-etal-2016-aspect}, IAN \citep{ma2017interactive}, AOA \citep{huang2018aspect}, AEN-BERT \citep{song2019attentional}, CDT \citep{sun2019aspect}, BERT+GCN, ASGCN and  ASGCN-BERT \citep{zhang2019aspect}, RGAT and RGAT-BERT \citep{wang2020relational}, and DGEDT-BERT \citep{tang-etal-2020-dependency}. We divide them into three categories. The category \textbf{Others} includes models that are general and task agnostic ones. The LSTM acts as a baseline for models without using PLM, while BERT-SPC and RoBERTa-MLP are baselines for models with the corresponding PLMs. The categories \textbf{Attention} and \textbf{Graph} are models mainly based on attention mechanism and graph neural networks, respectively. The details of datasets and experimental setup can be found in Appendix \ref{appendix:datasets}.

\subsection{Experimental Results}
\begin{table*}[htp]
\centering\small
\begin{adjustbox}{max width=2.0\textwidth}
\begin{tabular}{clcccccc}
\toprule
\multirow{2}{*}{Category} &
\multirow{2}{*}{Model}  &
\multicolumn{2}{c}{Twitter} &
\multicolumn{2}{c}{Laptop} & 
\multicolumn{2}{c}{Restaurant}  \\ 
\cmidrule{3-8}
& & Acc. & Macro-F1
& Acc. & Macro-F1
& Acc. & Macro-F1 \\
\midrule
\multirow{3}{*}{Others}&LSTM & 67.69 & 65.61 & 67.04 & 58.26 & 75.63 & 61.33 \\
& BERT-SPC & 72.88 & 71.78 & 76.19 & 70.84 & 83.74 & 74.77 \\
& RoBERTa-MLP & 72.76 & 71.73 & 81.11 & 77.11 & \textbf{86.79} & \textbf{79.76} \\
\midrule
\multirow{4}{*}{Attention} & MemNet & 68.41 & 65.63 & 68.35 & 61.61 & 77.68 & 64.99 \\
& IAN & 70.33 & 68.12 & 68.24 & 59.49 & 75.78 & 61.12 \\
& AOA & 69.61 & 67.07 & 69.61 & 62.84 & 77.12 & 63.73 \\
& AEN-BERT & 71.13 & 70.00 & 75.64 & 69.04 & 80.77 & 68.87 \\
\midrule
\multirow{5}{*}{Graph} & CDT & 71.93 & 69.96 & 72.38 & 67.16 & 79.90 & 68.83 \\
& BERT+GCN & 72.46 & 71.68 & 73.89 & 67.62 & 82.44 & 72.33 \\
& ASGCN & 69.28 & 66.63 & 70.55 & 64.26 & 78.65 & 67.03 \\
& ASGCN-BERT & 72.20 & 70.82 & 76.71 & 71.61 & 81.74 & 71.00 \\
& RGAT & 68.24 & 66.29 & 71.47 & 65.29 & 79.77 & 69.43 \\
& RGAT-BERT & 72.62 & 71.34 & 75.67 & 70.53 & 83.23 & 74.26  \\
& DGEDT-BERT & 72.79 & 71.35 & 76.03 & 70.70 & 83.85 & 75.18 \\
\midrule
\multirow{4}{*}{Ours} & BERT4GCN & \textbf{74.73} & \textbf{73.76} & \textbf{77.49} & \textbf{73.01} & \textbf{84.75} & \textbf{77.11} \\
& w/o pos.  & 74.48 & 73.22 & 77.24 & 72.82 & 84.41 & 76.29 \\
& w/o att.  & 74.15 & 72.90 & 77.12 & 72.46 & 83.91 & 75.28 \\
& w/o pos. \& w/o att. & 74.36 & 73.31 & 77.23 & 72.58 &  84.09 & 75.93 \\
& RoBERTa4GCN & \textbf{74.75} & \textbf{74.00} & \textbf{81.80} & \textbf{78.16} & 86.23 & 78.61 \\
\bottomrule
\end{tabular}
\end{adjustbox}
\caption{Comparisons of BERT4GCN with various baselines. The w/o pos. indicates the one without using relative position module and w/o att. is without using supplemented dependency graph. Accuracy (Acc.) and Marco-F1 are used for metrics. And we report the average of 10-fold experimental results (\%).}
\label{tab:results}
\end{table*}

\subsubsection{Overall Results}
We now present the comparisons of performance of BERT4GCN with other models in terms of classification accuracy and Macro-F1 on Table~\ref{tab:results}. 
From the table, we can observe that our BERT4GCN outperforms all other BERT-based models across all three datasets, justifying that augmenting GCN with the grammatical features from intermediate layers of BERT can empower ABSC models.

\subsubsection{Results Analysis}
\xhdr{BERT-SPC vs. RoBERTa-MLP} 
RoBERTa-MLP significantly outperforms BERT-SPC on Laptop and Restaurant datasets, while has similar results to BERT-SPC on Twitter dataset. The same pattern is also observed in the comparison of BERT4GCN and RoBERTa4GCN. One possible reason for this phenomenon is that the corpora which the two PLMs were pre-trained on are far different from Twitter dataset. Therefore, RoBERTa's superiority is not shown on the Twitter dataset, which we call the \textbf{out-domain dataset}.

\xhdr{BERT-SPC vs. BERT-based models}
From the table, we also see that BERT-SPC can parallel BERT-based models, indicating that model architecture engineering only has a marginal effect when using BERT.

\xhdr{BERT-SPC vs. BERT4GCN}
Observed from the experimental results, the improved performance of BERT4GCN  over BERT-SPC on Twitter dataset is higher than the other datasets. A similar pattern also appears in the comparison of RoBERTa4GCN and RoBERTa-MLP, where RoBERTa4GCN is just comparable to RoBERTa-MLP on Laptop and Restaurant datasets. For this phenomenon, we have two conjectures: (1) BERT4GCN framework is more flexible to handle the \textbf{out-domain dataset}; (2) When PLM is strong enough, with the current model framework, heavy model architecture engineering is unnecessary on the \textbf{in-domain dataset}. These two conjectures need to be further explored in future work.

\subsubsection{Ablation Study}
To examine the effectiveness of different modules in BERT4GCN, we have carried out ablation studies as shown in Table~\ref{tab:results}. 
As presented in Table~\ref{tab:results}, the full model of BERT4GCN has the best performance. And 
we observe that only fusing grammatical sequential features with GCN (i.e., w/o pos. \& w/o att.) can still achieve state-of-the-art results. With the supplemented the dependency graphs (i.e., w/o pos.), the performance increases just slightly.  

It is notable that adding the relative position module alone (w/o att) produces a negative effect, and the power of the relative position module can only be revealed when combined together with the supplemented dependency graphs. We conjecture that the supplemented dependency graphs prune some edges which connect nearby aspect terms to words that are irrelevant to sentiment classification, thus reducing the noise.

\begin{figure}[t]
\centering
\includegraphics[scale=0.6]{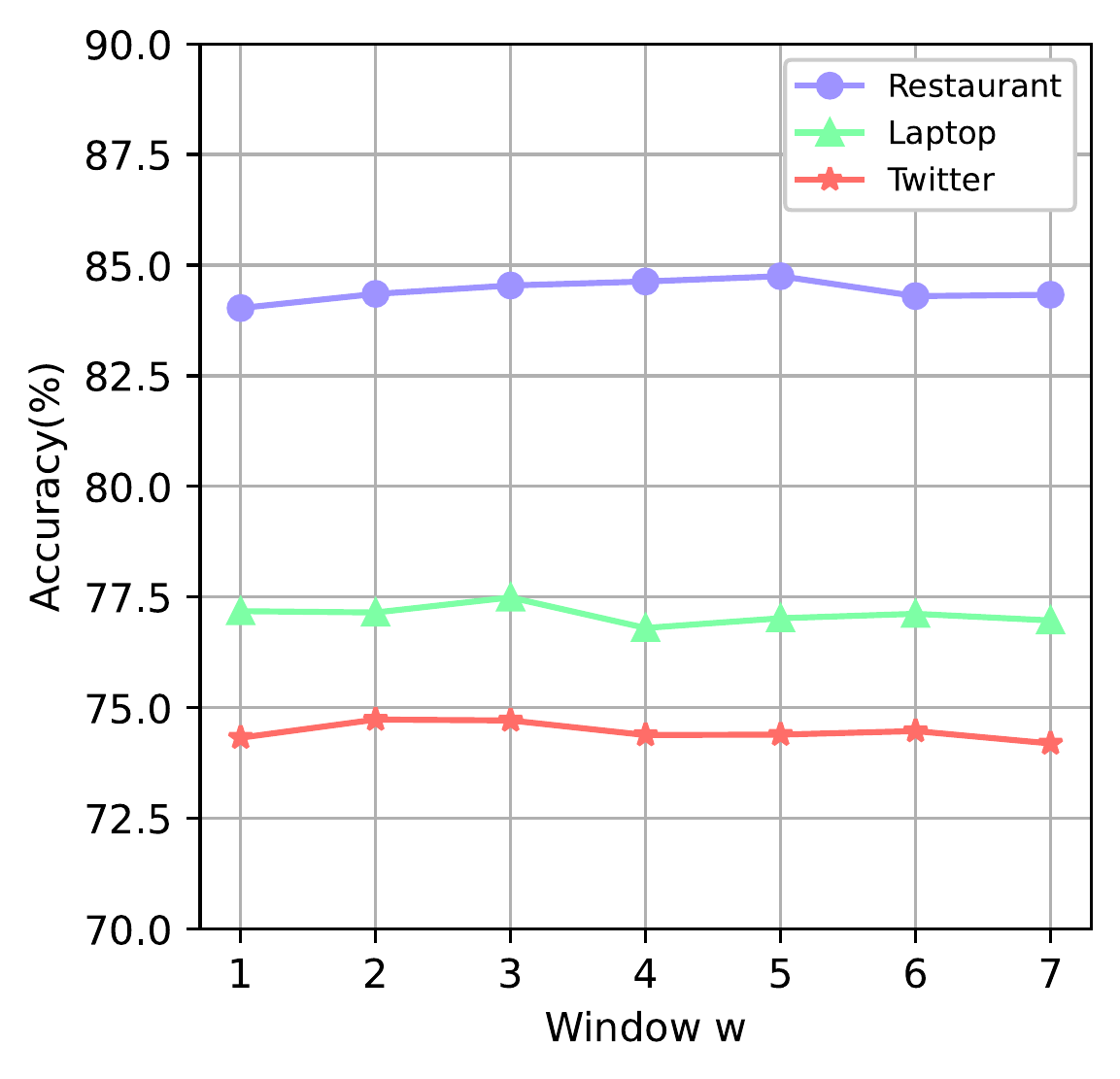}
\caption{Effect of the size of relative position window $w$ on Twitter (bottom), Laptop (mid) and Restaurant (top) datasets.}
\label{fig:win}
\end{figure}

\begin{figure}[t]
\centering
\includegraphics[width=1.0\columnwidth]{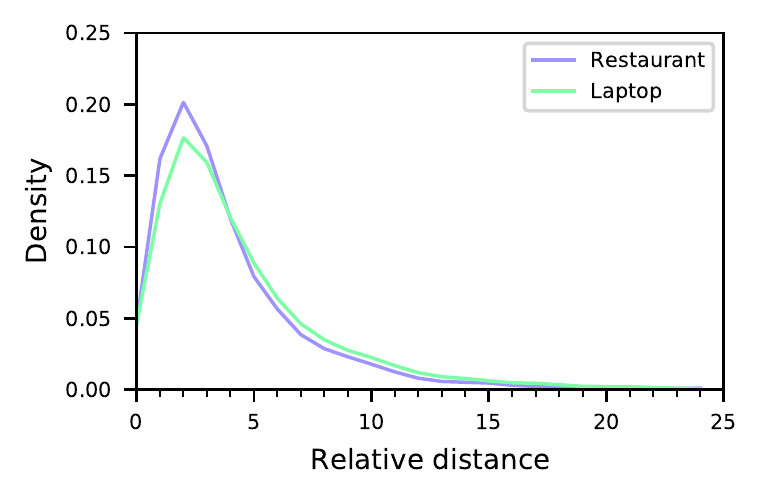}
\caption{Distributions of the distances between aspect and opinion terms on Laptop and Restaurant datasets.}
\label{fig:asp_op}
\end{figure}

\subsubsection{Effect of Relative Position Window}
We have also investigated the effect of the relative position window $w$ on BERT4GCN across three datasets. As shown in Figure~\ref{fig:win}, Twitter and Laptop datasets prefer a smaller window, while Restaurant dataset prefers a bigger one.

We calculate the relative distances between aspect and opinion terms of Laptop and Restaurant datasets with annotated aspect-opinion pairs by \citet{fan-etal-2019-target}, and visualize the distributions with kernel density estimation in Figure~\ref{fig:asp_op}. Although the two distributions are very similar, the optimal window size for the two datasets is not the same. Therefore we hypothesize that the preference of window size is also influenced by the dataset domain.
The long-tailed distributions imply that we need to carefully set the window size for the trade-off between the benefits and losses of position biases.

\section{Conclusion}
In this paper, we propose a BERT4GCN model which integrates the grammatical sequential features from BERT along with the syntactic knowledge from dependency graphs. The proposed model utilizes intermediate layers of BERT, which contain rich and helpful linguistic knowledge, to augment GCN, and furthermore,   incorporates relative positional information of words to be position-aware. Finally, experimental results show that our model achieves new state-of-the-art performance on prestigious benchmarks.

\section*{Acknowledgements}
This research is supported by Qinghai Provincial Science and Technology Research Program (grant No.2021-QY-206), Guangdong Provincial Science and Technology Research Program (grant No.2020A050515014), and National Natural Science Foundation of China (grant No. 62071201).



\begin{thebibliography}{22}
\expandafter\ifx\csname natexlab\endcsname\relax\def\natexlab#1{#1}\fi

\bibitem[{Clark et~al.(2019)Clark, Khandelwal, Levy, and
  Manning}]{clark2019does}
Kevin Clark, Urvashi Khandelwal, Omer Levy, and Christopher~D. Manning. 2019.
\newblock \href {https://doi.org/10.18653/v1/W19-4828} {What does {BERT} look
  at? an analysis of {BERT}{'}s attention}.
\newblock In \emph{Proceedings of the 2019 ACL Workshop BlackboxNLP: Analyzing
  and Interpreting Neural Networks for NLP}, pages 276--286, Florence, Italy.
  Association for Computational Linguistics.

\bibitem[{Dai et~al.(2021)Dai, Yan, Sun, Liu, and Qiu}]{dai2021does}
Junqi Dai, Hang Yan, Tianxiang Sun, Pengfei Liu, and Xipeng Qiu. 2021.
\newblock Does syntax matter? a strong baseline for aspect-based sentiment
  analysis with roberta.
\newblock In \emph{Proceedings of the 2021 Conference of the North American
  Chapter of the Association for Computational Linguistics: Human Language
  Technologies}, pages 1816--1829.

\bibitem[{Devlin et~al.(2019)Devlin, Chang, Lee, and
  Toutanova}]{devlin2018bert}
Jacob Devlin, Ming-Wei Chang, Kenton Lee, and Kristina Toutanova. 2019.
\newblock \href {https://doi.org/10.18653/v1/N19-1423} {{BERT}: Pre-training of
  deep bidirectional transformers for language understanding}.
\newblock In \emph{Proceedings of the 2019 Conference of the North {A}merican
  Chapter of the Association for Computational Linguistics: Human Language
  Technologies, Volume 1 (Long and Short Papers)}, pages 4171--4186,
  Minneapolis, Minnesota. Association for Computational Linguistics.

\bibitem[{Dong et~al.(2014)Dong, Wei, Tan, Tang, Zhou, and
  Xu}]{dong-etal-2014-adaptive}
Li~Dong, Furu Wei, Chuanqi Tan, Duyu Tang, Ming Zhou, and Ke~Xu. 2014.
\newblock \href {https://doi.org/10.3115/v1/P14-2009} {Adaptive recursive
  neural network for target-dependent {T}witter sentiment classification}.
\newblock In \emph{Proceedings of the 52nd Annual Meeting of the Association
  for Computational Linguistics (Volume 2: Short Papers)}, pages 49--54,
  Baltimore, Maryland. Association for Computational Linguistics.

\bibitem[{Fan et~al.(2019)Fan, Wu, Dai, Huang, and Chen}]{fan-etal-2019-target}
Zhifang Fan, Zhen Wu, Xin-Yu Dai, Shujian Huang, and Jiajun Chen. 2019.
\newblock \href {https://doi.org/10.18653/v1/N19-1259} {Target-oriented opinion
  words extraction with target-fused neural sequence labeling}.
\newblock In \emph{Proceedings of the 2019 Conference of the North {A}merican
  Chapter of the Association for Computational Linguistics: Human Language
  Technologies, Volume 1 (Long and Short Papers)}, pages 2509--2518,
  Minneapolis, Minnesota. Association for Computational Linguistics.

\bibitem[{Hewitt and Liang(2019)}]{hewitt-liang-2019-designing}
John Hewitt and Percy Liang. 2019.
\newblock \href {https://doi.org/10.18653/v1/D19-1275} {Designing and
  interpreting probes with control tasks}.
\newblock In \emph{Proceedings of the 2019 Conference on Empirical Methods in
  Natural Language Processing and the 9th International Joint Conference on
  Natural Language Processing (EMNLP-IJCNLP)}, pages 2733--2743, Hong Kong,
  China. Association for Computational Linguistics.

\bibitem[{Hewitt and Manning(2019)}]{hewitt-manning-2019-structural}
John Hewitt and Christopher~D. Manning. 2019.
\newblock \href {https://doi.org/10.18653/v1/N19-1419} {{A} structural probe
  for finding syntax in word representations}.
\newblock In \emph{Proceedings of the 2019 Conference of the North {A}merican
  Chapter of the Association for Computational Linguistics: Human Language
  Technologies, Volume 1 (Long and Short Papers)}, pages 4129--4138,
  Minneapolis, Minnesota. Association for Computational Linguistics.

\bibitem[{Huang and Carley(2019)}]{huang2019syntax}
Binxuan Huang and Kathleen Carley. 2019.
\newblock \href {https://doi.org/10.18653/v1/D19-1549} {Syntax-aware aspect
  level sentiment classification with graph attention networks}.
\newblock In \emph{Proceedings of the 2019 Conference on Empirical Methods in
  Natural Language Processing and the 9th International Joint Conference on
  Natural Language Processing (EMNLP-IJCNLP)}, pages 5469--5477, Hong Kong,
  China. Association for Computational Linguistics.

\bibitem[{Huang et~al.(2018)Huang, Ou, and Carley}]{huang2018aspect}
Binxuan Huang, Yanglan Ou, and Kathleen~M Carley. 2018.
\newblock Aspect level sentiment classification with attention-over-attention
  neural networks.
\newblock In \emph{International Conference on Social Computing,
  Behavioral-Cultural Modeling and Prediction and Behavior Representation in
  Modeling and Simulation}, pages 197--206. Springer.

\bibitem[{Huang et~al.(2020)Huang, Sun, Li, Zhang, and Wang}]{huang2020syntax}
Lianzhe Huang, Xin Sun, Sujian Li, Linhao Zhang, and Houfeng Wang. 2020.
\newblock Syntax-aware graph attention network for aspect-level sentiment
  classification.
\newblock In \emph{Proceedings of the 28th International Conference on
  Computational Linguistics}, pages 799--810.

\bibitem[{Jawahar et~al.(2019)Jawahar, Sagot, and
  Seddah}]{jawahar-etal-2019-bert}
Ganesh Jawahar, Beno{\^\i}t Sagot, and Djam{\'e} Seddah. 2019.
\newblock \href {https://doi.org/10.18653/v1/P19-1356} {What does {BERT} learn
  about the structure of language?}
\newblock In \emph{Proceedings of the 57th Annual Meeting of the Association
  for Computational Linguistics}, pages 3651--3657, Florence, Italy.
  Association for Computational Linguistics.

\bibitem[{Kipf and Welling(2017)}]{kipf2016semi}
Thomas~N. Kipf and Max Welling. 2017.
\newblock \href {https://openreview.net/forum?id=SJU4ayYgl} {Semi-supervised
  classification with graph convolutional networks}.
\newblock In \emph{5th International Conference on Learning Representations,
  {ICLR} 2017, Toulon, France, April 24-26, 2017, Conference Track
  Proceedings}. OpenReview.net.

\bibitem[{Ma et~al.(2017)Ma, Li, Zhang, and Wang}]{ma2017interactive}
Dehong Ma, Sujian Li, Xiaodong Zhang, and Houfeng Wang. 2017.
\newblock \href {https://doi.org/10.24963/ijcai.2017/568} {Interactive
  attention networks for aspect-level sentiment classification}.
\newblock In \emph{Proceedings of the Twenty-Sixth International Joint
  Conference on Artificial Intelligence, {IJCAI} 2017, Melbourne, Australia,
  August 19-25, 2017}, pages 4068--4074. ijcai.org.

\bibitem[{Pontiki et~al.(2014)Pontiki, Galanis, Pavlopoulos, Papageorgiou,
  Androutsopoulos, and Manandhar}]{pontiki-etal-2014-semeval}
Maria Pontiki, Dimitris Galanis, John Pavlopoulos, Harris Papageorgiou, Ion
  Androutsopoulos, and Suresh Manandhar. 2014.
\newblock \href {https://doi.org/10.3115/v1/S14-2004} {{S}em{E}val-2014 task 4:
  Aspect based sentiment analysis}.
\newblock In \emph{Proceedings of the 8th International Workshop on Semantic
  Evaluation ({S}em{E}val 2014)}, pages 27--35, Dublin, Ireland. Association
  for Computational Linguistics.

\bibitem[{Song et~al.(2019)Song, Wang, Jiang, Liu, and
  Rao}]{song2019attentional}
Youwei Song, Jiahai Wang, Tao Jiang, Zhiyue Liu, and Yanghui Rao. 2019.
\newblock Attentional encoder network for targeted sentiment classification.
\newblock \emph{arXiv preprint arXiv:1902.09314}.

\bibitem[{Sun et~al.(2019)Sun, Zhang, Mensah, Mao, and Liu}]{sun2019aspect}
Kai Sun, Richong Zhang, Samuel Mensah, Yongyi Mao, and Xudong Liu. 2019.
\newblock \href {https://doi.org/10.18653/v1/D19-1569} {Aspect-level sentiment
  analysis via convolution over dependency tree}.
\newblock In \emph{Proceedings of the 2019 Conference on Empirical Methods in
  Natural Language Processing and the 9th International Joint Conference on
  Natural Language Processing (EMNLP-IJCNLP)}, pages 5679--5688, Hong Kong,
  China. Association for Computational Linguistics.

\bibitem[{Tang et~al.(2016)Tang, Qin, and Liu}]{tang-etal-2016-aspect}
Duyu Tang, Bing Qin, and Ting Liu. 2016.
\newblock \href {https://doi.org/10.18653/v1/D16-1021} {Aspect level sentiment
  classification with deep memory network}.
\newblock In \emph{Proceedings of the 2016 Conference on Empirical Methods in
  Natural Language Processing}, pages 214--224, Austin, Texas. Association for
  Computational Linguistics.

\bibitem[{Tang et~al.(2020)Tang, Ji, Li, and Zhou}]{tang-etal-2020-dependency}
Hao Tang, Donghong Ji, Chenliang Li, and Qiji Zhou. 2020.
\newblock \href {https://doi.org/10.18653/v1/2020.acl-main.588} {Dependency
  graph enhanced dual-transformer structure for aspect-based sentiment
  classification}.
\newblock In \emph{Proceedings of the 58th Annual Meeting of the Association
  for Computational Linguistics}, pages 6578--6588, Online. Association for
  Computational Linguistics.

\bibitem[{Vaswani et~al.(2017)Vaswani, Shazeer, Parmar, Uszkoreit, Jones,
  Gomez, Kaiser, and Polosukhin}]{vaswani2017attention}
Ashish Vaswani, Noam Shazeer, Niki Parmar, Jakob Uszkoreit, Llion Jones,
  Aidan~N Gomez, {\L}ukasz Kaiser, and Illia Polosukhin. 2017.
\newblock Attention is all you need.
\newblock In \emph{Advances in neural information processing systems}, pages
  5998--6008.

\bibitem[{Velickovic et~al.(2018)Velickovic, Cucurull, Casanova, Romero,
  Li{\`{o}}, and Bengio}]{velivckovic2017graph}
Petar Velickovic, Guillem Cucurull, Arantxa Casanova, Adriana Romero, Pietro
  Li{\`{o}}, and Yoshua Bengio. 2018.
\newblock \href {https://openreview.net/forum?id=rJXMpikCZ} {Graph attention
  networks}.
\newblock In \emph{6th International Conference on Learning Representations,
  {ICLR} 2018, Vancouver, BC, Canada, April 30 - May 3, 2018, Conference Track
  Proceedings}. OpenReview.net.

\bibitem[{Wang et~al.(2020)Wang, Shen, Yang, Quan, and
  Wang}]{wang2020relational}
Kai Wang, Weizhou Shen, Yunyi Yang, Xiaojun Quan, and Rui Wang. 2020.
\newblock \href {https://doi.org/10.18653/v1/2020.acl-main.295} {Relational
  graph attention network for aspect-based sentiment analysis}.
\newblock In \emph{Proceedings of the 58th Annual Meeting of the Association
  for Computational Linguistics}, pages 3229--3238, Online. Association for
  Computational Linguistics.

\bibitem[{Zhang et~al.(2019)Zhang, Li, and Song}]{zhang2019aspect}
Chen Zhang, Qiuchi Li, and Dawei Song. 2019.
\newblock \href {https://doi.org/10.18653/v1/D19-1464} {Aspect-based sentiment
  classification with aspect-specific graph convolutional networks}.
\newblock In \emph{Proceedings of the 2019 Conference on Empirical Methods in
  Natural Language Processing and the 9th International Joint Conference on
  Natural Language Processing (EMNLP-IJCNLP)}, pages 4568--4578, Hong Kong,
  China. Association for Computational Linguistics.

\end{thebibliography}
\bibliographystyle{acl_natbib}

\clearpage
\appendix
\section{Appendix}
\label{sec:appendix}

\subsection{Datasets and Experimental Setup}
\label{appendix:datasets}
We have evaluated our BERT4GCN model on three widely used datasets, including Laptop and Restaurant datasets from SemEval 2014 Task 4 and Twitter datasets. The statistics of these three datasets are listed in Table~\ref{tab:datasets}. Just like other research work, we remove samples with conflicting polarities.
\begin{table}[h]
\centering
\begin{adjustbox}{max width=0.48\textwidth}
\begin{tabular}{ c c c c c c c }
\hline
\multirow{2}{*}{Dataset} & \multicolumn{2}{c}{Positive} & \multicolumn{2}{c}{Neutral} & \multicolumn{2}{c}{Negative} \\
\cline{2-7}
& Train & Test & Train & Test & Train & Test \\
\hline
Twitter & 1561 & 173 & 3127 & 346 & 1560 & 173  \\
Laptop & 994 & 341 & 464 & 169 & 870 & 128 \\
Restaurant & 2164 & 728 & 637 & 196 & 807 & 196 \\
\hline
\end{tabular}
\end{adjustbox}
\caption{Statistics of datasets}
\label{tab:datasets}
\end{table}

We use $10$-fold cross validation to evaluate the model performance. Specifically, we evaluate models at the end of epochs and save the checkpoint that achieves highest accuracy in the validation set. Finally, we test the models on the test set and report the average of 10-fold experimental results.

\subsection{Implementation Details}
\label{appendix:implementation}
We choose 300-dimensional Glove vectors for the word embeddings.
During training, the learning rate of BERT and other modules linearly warms up from $0$ to $0.00002$ and $0.001$, respectively, then linearly decreases to 0.  

In this paper, we name our model as BERT4GCN and use the bert-base-uncased model as the source of grammatical sequential features, but it is easy to extend it to other transformer-based pre-training language models\footnote{For RoBERTa4GCN, we use roberta-base model.}. 

The training of BERT4GCN model has been run on a Nvidia RTX 3090 that requires about 14GB of GPU memory. The PyTorch implementation of BERT\footnote{https://github.com/huggingface/transformers} is used in the experiments. 
The dimensionality of unidirectional LSTM hidden state and positional embedding is set to 300.
We set $\alpha$ to 0.01 and $\beta $ to 0.25, respectively.
The position window $w$ is set to 2, 3, 5 on Twitter, Laptop and Restaurant datasets, respectively.
And the batch size is set to $32$, with Adam as the optimizer.
As for the regularization, dropout is applied to BERT and GCN output with the rate of $0.8$.
And we use Spacy\footnote{https://spacy.io/} toolkit to generate dependency trees.

For comparisons of baseline models, we use their official implementations or the implementation in ABSA-PyTorch \footnote{https://github.com/songyouwei/ABSA-PyTorch}, configured with recommended hyperparameters. For BERT+GCN model, we build the graph in the same way as \citet{huang2020syntax}.

\end{document}